\documentclass[conference]{IEEEtran}
\IEEEoverridecommandlockouts
\usepackage{cite}
\usepackage{amsmath,amssymb,amsfonts}
\usepackage{graphicx}
\usepackage{multirow}
\usepackage{booktabs}
\usepackage{color, soul}
\newcommand{\blue}[1]{\textcolor{blue}{#1}}

\usepackage{algorithm}
\usepackage{algorithmic}
\usepackage{textcomp}
\def\BibTeX{{\rm B\kern-.05em{\sc i\kern-.025em b}\kern-.08em
    T\kern-.1667em\lower.7ex\hbox{E}\kern-.125emX}}
\begin{document}

\title{Efficient Diversity-Driven Ensemble for Deep Neural Networks
}

\author{
\IEEEauthorblockN{
$\sharp$Wentao Zhang~~~~~$\dag$Jiawei Jiang$^*$\thanks{* Jiawei Jiang is the corresponding author}~~~~~~$\ddag$Yingxia Shao~~~~~~$\sharp\P$Bin Cui}
\IEEEauthorblockA{\textit{$\sharp$Center for Data Science, Peking University \& National Engineering Laboratory for Big Data Analysis and Applications} \\
\textit{$\P$School of EECS \& Key Laboratory of High Confidence Software Technologies (MOE), Peking University} \\
\textit{$\dag$ETH Zurich, Switzerland~~~$\ddag$School of Computer Science, Beijing University of Posts and Telecommunications.}\\
\textit{$\P$\{wentaozhang, bin.cui\}@pku.edu.cn, $\dag$jiawei.jiang@inf.ethz.ch, $\ddag$shaoyx@bupt.edu.cn}\\} \\
}

\maketitle

\begin{abstract}

The ensemble of deep neural networks has been shown, both theoretically and empirically, to improve generalization accuracy on the unseen test set. However, the high training cost hinders its efficiency since we need a sufficient number of base models and each one in the ensemble has to be separately trained.
Lots of methods are proposed to tackle this problem, and most of them are based on the feature that a pre-trained network can transfer its knowledge to the next base model and then accelerate the training process.
However, these methods suffer a severe problem that
all of them transfer knowledge without selection and thus lead to low diversity. 
As the effect of ensemble learning is more pronounced if ensemble members are accurate and diverse, 
we propose a method named Efficient Diversity-Driven Ensemble (EDDE) to address both the diversity and the efficiency of an ensemble. 
To accelerate the training process, we propose a novel knowledge transfer method which can selectively transfer the previous generic knowledge. 
To enhance diversity, we first propose a new diversity measure, then use it to define a diversity-driven loss function for optimization. 
At last, we adopt a Boosting-based framework to combine the above operations, such a method can also further improve diversity. We evaluate EDDE on Computer Vision (CV) and Natural Language Processing (NLP) tasks. Compared with other well-known ensemble methods, EDDE can get highest ensemble accuracy with the lowest training cost, which means it is efficient in the ensemble of neural networks.

\end{abstract}

\begin{IEEEkeywords}
deep neural networks, ensemble learning, knowledge transfer, diversity, efficient
\end{IEEEkeywords}

\section{Introduction}
Deep Neural Network has aroused people's concern and it has been widely used in many real applications~\cite{ciresan2011flexible,simon}.
Unfortunately, a neural network may converge to different local minimums during the training process, and its generalization ability may be unstable accordingly~\cite{zhang2016understanding}. Lots of methods have been proposed to tackle this problem, one famous approach is ensemble learning.

The ensemble of deep neural networks trained on the same dataset is proved theoretically and experimentally to improve the generalization accuracy in many practical applications\cite{time-ens,wind}. 
For example, motivated by the idea of Knowledge Distillation~\cite{hinton2015distilling}, the base models in Born-Again Networks(BANs) are trained from the supervision of the earlier fitted model~\cite{BANs}.
By training with the objective of matching the full softmax distribution of the pre-trained model, BANs can get a rich source of training signal. 
As a result, additional gains can be achieved with an ensemble of multiple base models.

Like BANs, ensemble learning is usually designed to randomly initialize each base model. Accordingly, the neural network can converge to different local minimums and tend to make different predictions on the same sample, then the weak samples can benefit from the ensemble of these neural networks~\cite{dietterich2002ensemble}.
However, the training cost of the ensemble grows linearly with the number of the base models since each neural network added to the ensemble needs to be trained. Even on high-performance hardware, training a single deep neural network can take several days to complete. If each base model is trained from scratch, a large ensemble of neural networks will require significant training time and computational resources.

To reduce the training cost, many useful methods have been proposed in recent years\cite{LF,mosca2017deep,Snap-Boost}. A commonly used idea is to transfer the knowledge of the pre-trained model to next base model, and the most typical one is Snapshot Ensemble~\cite{huang2017snapshot}.
By periodically resetting the learning rate and decaying it with a cosine annealing~\cite{sgdr}, a neural network in Snapshot Ensemble can converge to different local minimums along its optimization path. Unlike Bagging~\cite{breiman1996stacked}, which trains each individual network independently, Snapshot Ensemble saves the model parameter before resetting the learning rate and treats each model replica as a base model. Therefore, it has the ability to get multiple base models with less training cost.


However, the neural network has many local minimums, and the base models in an ensemble may be trapped in the same or nearby local minimums. We call this phenomenon low diversity, which causes less accurate ensemble model. Specifically, this phenomenon is more likely to happen in Snapshot Ensemble. 
Initialized with the weights of the former base model, the current neural network in Snapshot Ensemble is unable to learn more specific knowledge if given limited training budget, thus it would make similar predictions to the former one and the diversity is reduced accordingly. 
The ensemble accuracy can be significantly improved if the base models have both high diversity and accuracy~\cite{polikar2006ensemble}. As a result, Snapshot Ensemble is not efficient to get high generalization accuracy due to its low diversity.


To enhance diversity, the goal is to get multiple base models whose predictions are negatively correlated. Negative Correlation Learning (NCL)~\cite{liu1999ensemble} is proposed to achieve this goal, it negatively correlates the errors made by each base model explicitly so that the diversity can be enhanced. Based on this idea, AdaBoost.NC~\cite{wang2010negative} uses an ambiguity term derived theoretically for classification ensemble to introduce diversity. Experiments show it can get both high diversity and generalization accuracy than the original AdaBoost.M2~\cite{rokach2010ensemble} algorithm.
However, AdaBoost.NC studies the ambiguity decomposition with the 0-1 error function, such diversity definition is flawed as it loses massive raw information of the softmax outputs. Besides, AdaBoost.NC trains each base model without using the prior knowledge. So, the base model in AdaBoost.NC cannot converge well in a limited time and thus this will lead to an ensemble with  high bias.

To sum up, all the methods above have non-negligible drawbacks regarding the training cost or diversity, and there 
lacks an ensemble learning method which is able to train diverse and accurate neural networks with a limited training budget.
Therefore,
we study how to reduce the training cost and increase the diversity of an ensemble.

To accelerate convergence, we propose to adaptively transfer knowledge contained in one neural network to another neural network. It is well established that DNN generally learns generic features in its lower layers and task-specific features in its upper layers~\cite{trans}. As shown in Snapshot Ensemble, it transfers both the generic and task-specific features, which results in low diversity. To avoid this situation, we propose a method to efficiently select the generic knowledge contained in the first several layers of a neural network. 
By transferring this previous generic knowledge, we can train each neural network significantly faster than approaches that train each model from scratch. More importantly, the total training time will grow 
much slower as we increase the ensemble size. 

To guarantee diversity, we propose a diversity-driven loss function to optimize each base model. 
This loss function has a weighted average of two different objective functions. The first one is to learn the distribution from the training data, and the other is to negatively correlate the softmax outputs which also called soft target~\cite{hinton2015distilling} of the former ensemble network. For the classification tasks using deep neural networks, our work is the first, to the best of our knowledge, to achieve the diversity explicitly by using a diversity-driven loss function.

\begin{figure}[htbp]
\centering
\includegraphics[width=5.5cm]{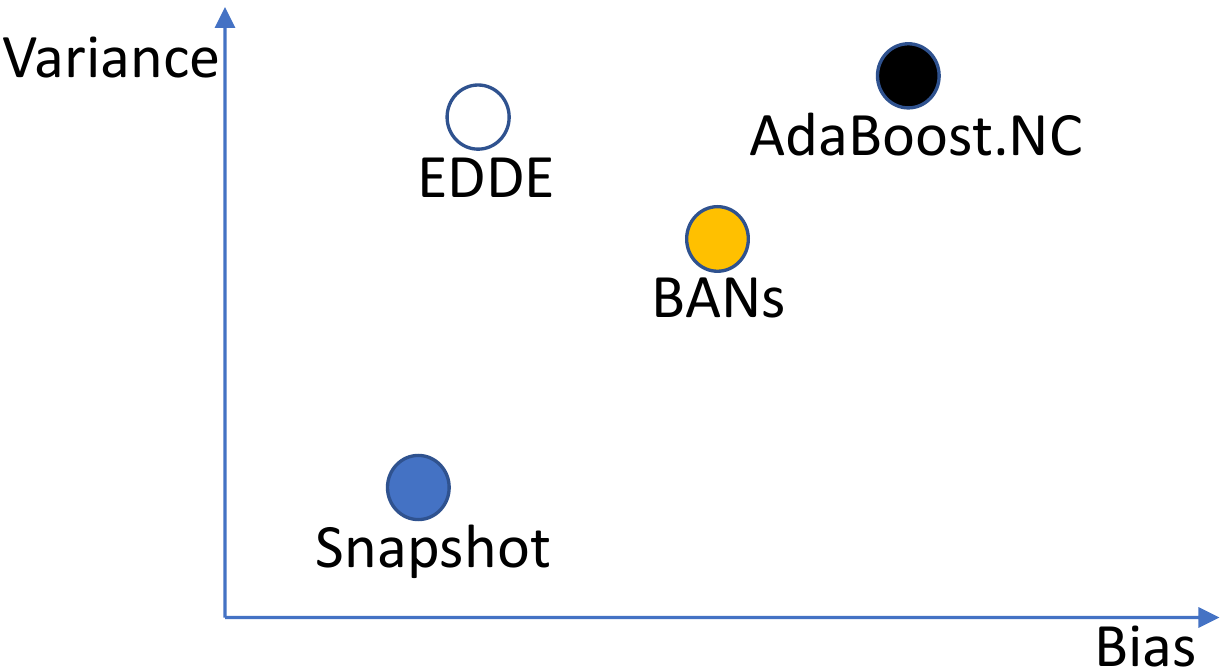}
\caption{The bias and variance analysis of each method.}
\label{BiaVar}
\vspace{-15pt}
\end{figure}

Last, as Boosting is a  diversity encouraging strategy, we use Boosting to construct the training pipeline for further improving the diversity. Our method is efficient in producing high diversity base models, thus we call it Efficient Diversity-Driven Ensemble(EDDE). To theoretically analyze the relationship between EDDE and other representative methods, we calculate their bias and variance within the same training budget. 
We run this experiment on the dataset CIFAR100~\cite{Cifar} using ResNet-32~\cite{he2016deep}, and the result is shown in Figure~\ref{BiaVar}. 

In theory, the ensemble accuracy can be significantly improved if the base models in the ensemble are accurate and diverse~\cite{grani}, which is equal to say the base models of a good ensemble should both have low bias and high variance. 
Seen from Figure~\ref{BiaVar},
given the same and limited training budget, AdaBoost.NC can get the highest variance but also introduce the highest bias.
On the contrary, Snapshot Ensemble is able to train a set of base models with low bias but it also faces a problem of low variance.
As for BANs, it cannot get a high variance and low bias at the same time.
Therefore, all these baselines face the bias-variance dilemma~\cite{geman}.
Our method EDDE outperforms them since it can rapidly produce a set of base models which has low bias and high variance.  Given a limited training budget, we test EDDE in the CV and NLP scenarios, and the experiments show that it can get the state-of-art ensemble accuracy than other well-known methods.

The main contributions of EDDE can be summarized as follows:
\begin{itemize}
    \item We propose a knowledge transfer strategy which can accelerate the training process without reducing the diversity and we also propose a method to efficiently select the generic knowledge which we should transfer.
    
    \item We propose a novel diversity measure and use it to explicitly define a diversity-driven loss function. Besides, such a measure can numerically compare the diversity among each ensemble method.
    
    \item We propose a Boosting-based framework to combine the knowledge transfer and optimization process, and such training pipeline can further improve the diversity.
    
\end{itemize}

The remainder of this paper is organized as follows. We describe the related work in Section II. Section III introduces some preliminaries and notations. The learning algorithm is introduced in Section IV and Section V presents the experimental results. Finally, we conclude this paper in Section VI.

\section{Related work}
\subsection{Ensemble learning}
Ensemble learning is an art of combining the predictions of different machine learning models, and it has achieved state-of-the-art performance in many challenges such as the Netflix Prize and various Kaggle competitions~\cite{hoch2015ensemble}. From the algorithmic level, Bagging, Boosting and Stacking are very representative algorithms.

Bagging is an ensemble strategy which uses the bootstrap aggregation to train multiple models while reducing variance. As a popular bagging method,  Majority Voting~\cite{dietterich2000ensemble} counts the votes from all base learners and makes the prediction using the most voted label. The other method is Averaging~\cite{Ury1997Optimal}, which averages the softmax outputs of base learners. By averaging six residual networks, He et al~\cite{he2016deep} won the first place in ILSVRC 2015. Similarly, Snapshot Ensemble and BANs adopts an average strategy in its prediction process.  

Another popular ensemble method is Boosting.
It trains multiple base models sequentially and re-weights the training samples which are harder to train so that the latter models focus more on the difficult examples~\cite{Hastie2010The}.
The idea of boosting neural networks has been around for many years~\cite{druc}. BoostCNN incorporates boosting weights into the deep learning architecture~\cite{BoostedCNN}, and it can select the best network structure in each iteration. Besides, deep incremental boosting~\cite{mosca2017deep} uses transfer learning to accelerate the initial warm-up phase of training each Boosting round. To enhance diversity, AdaBoost.NC propose an ambiguity decomposition strategy to negatively correlate the error made by other models.

Unlike Bagging and Boosting, Stacking~\cite{breiman1996stacked} combines the outputs of each base model via a meta-learner. Based on the same idea, deep super learner~\cite{young2018deep} circularly appends the outputs of base models to the training data and uses the super learner to combine their outputs. Cheng Ju~\cite{ju2018relative} adopts super Learner from the convolution neural network perspective, in which a CNN network is used to combine each output of the base model.

Besides, according to the usage of the training data, current methods of training ensemble of the deep neural networks can be classified into two categories. The first one is to train each network architecture with the entire dataset~\cite{huang2017snapshot,why}, and the other one is to train it with a different subset of replacements from the original dataset~\cite{BoostedCNN}. 
Usually, a deep neural network has a large parameter space and it can get low bias if trained with the entire dataset~\cite{why}. However, this may bring higher training cost. As a sub-sampling method, Bagging and Boosting can enhance diversity but result in higher bias of base models since we train them with less unique samples in each iteration.

As we know, 
base models in an ensemble contribute more if they are diverse,
otherwise there may be no gain in the combining process. However, all these methods individually train each base model without correlation and thus they cannot introduce diversity in an explicit way. 
Since there is no feedback from the combination stage to the individual design stage, it is possible that some of the independently designed networks may not make any contribution to the whole ensemble model. Furthermore, some of them train each network from scratch, requiring a lot of time and training resource.

\subsection{Negative correlated learning}
Current methods for designing neural network ensemble usually train individual neural networks independently. However, the base models are likely to be trapped in the same nearby local minimums and result in low diversity, especially if we transfer the knowledge of the pre-trained base model~\cite{huang2017snapshot}. Both theoretical and experimental studies show that the ensemble generalization ability can be largely enhanced if the base models in the ensemble are negatively correlated~\cite{error}.

Negative correlation learning(NCL) is proposed to achieve this goal, all the individual networks in the NCL are trained simultaneously through the correlation penalty terms in their loss functions, thus the diversity can be introduced in an explicit way.

Based on the NCL, cooperative ensemble learning system (CELS)~\cite{liu1998cooperative} encourages the individual networks to learn different aspects of a given dataset cooperatively and simultaneously. Besides, learning via Correlation-Corrected Data(NCCD)~\cite{chan2005preliminary} embeds penalty values to every training sample instead of the error function. 

However, all of their penalty terms are unfit to implement in classification ensembles. The AdaBoost.NC is proposed to solve this problem, it defines the penalty term with the correct/incorrect decision and introduces the error correlation information into the weights of the training data. 

Although this definition can be used for classification ensembles with NCL, such coarse-grained definition may lose lots of useful information the model has learned before. Besides, AdaBoost.NC only provides a diversity-aware heuristic to adjust the weight of training data but fails to formalize an objective function that
takes diversity into consideration. Therefore, the diversity of base models in AdaBoost.NC can be inferior, which limits the ensemble performance.

Our method EDDE differs AdaBoost.NC in two aspects. First, we propose a new diversity-driven loss function to explicitly increase diversity, while AdaBoost.NC introduce diversity by adjusting the sample weights. Compared with our diversity-driven loss function, using the sample weights to enhance diversity is not direct and robust. Besides, EDDE  captures more information as we negatively correlate the soft target of former ensemble model.



\begin{table}[!htbp]
\caption{NOTATIONS}
\centering
{
\noindent
\renewcommand{\multirowsetup}{\centering}
\begin{tabular}{cc}
\toprule
\textbf{Symbols}& 
\textbf{Definitions}\\
\midrule
$D$& Training set \\
\midrule
$x_i$& The feature of the $i$-th training sample\\
\midrule
$y_i$& The label of the $i$-th training sample  \\
\midrule
$\boldsymbol{y_i}$& The one hot encoding of the $i$-th training label   \\
\midrule
$k$& The class of the training sample  \\
\midrule
$W$& The weights of training sample  \\
\midrule
$N$& The size of the training set  \\
\midrule
$T$& The number of training iterations  \\
\midrule
$\gamma$& Controls the strength of the diversity-driven loss  \\
\midrule
$\beta$& Controls the proportion of knowledge transfer \\
\midrule
$h_t$& The $t$-th base neural network\\
\midrule
$H_t$& The ensemble of $t$ base models\\
\midrule
$\alpha_t$& The weight of the $t$-th base model\\
\midrule
$h_{t,c}(x_i)$ & The $c$-th value of $h_t$'s soft target on sample $x_i$\\
\midrule
$h_{t}(x_i)$& The $t$-th base model's label prediction on sample $x_i$\\
\midrule
$\boldsymbol{h_{t}(x_i)}$& The $t$-th base model's soft target on sample $x_i$\\
\midrule
$\boldsymbol{H_{t}(x_i)}$& The $t$-th ensemble model's soft target on sample $x_i$\\
\midrule
$Sim_{t}(x_i)$& The similarity between model $h_t$ and $H_t$ on sample $x_i$\\
\midrule
$Bias_{t}(x_i)$& The bias of model $h_t$ on sample $x_i$\\
\bottomrule
\end{tabular}}
 \label{Notation}
\end{table}

\section{PRELIMINARIES}
In this section, we begin by introducing some notations related to EDDE.  Table~\ref{Notation} lists the symbols used in this paper.

  

The $i$-th sample in training set $D$ consists of the feature denoted by $x_i$ and the label denoted by $y_i$, the bold entity $\boldsymbol{y_i}$ denotes the one-hot encoding of the $i$-th training label and the number of classes is denoted by $k$.
Given $D$ with sample weights $W$ and sample size $N$, we can train a base model $h_t$. 

We use a diversity-driven optimization to enhance diversity, and $\gamma$ controls the strength of the diversity-driven loss.
Given a sample $x_i$, both the base model $h_t$ and the ensemble $H_t$ can make a prediction, thus we can accordingly get their softmax outputs as the soft-target $\boldsymbol{h_{t}(x_i)}$ and $\boldsymbol{H_{t}(x_i)}$. 
Note that ${h_{t}(x_i)}$ is the prediction label and the bold $\boldsymbol{h_{t}(x_i)}$ is a vector of probabilities representing the conditional distribution over object categories on sample $x_i$.
We set the $c$-th value of $\boldsymbol{h_{t}(x_i)}$ as $h_{t,c}(x_i)$, and this means the probability that sample $x_i$ belongs to the class $c$.
Besides, we define $Sim_{t}(x_i)$ as the similarity between model $h_t$ and $H_t$ on sample $x_i$, and we also set $Bias_{t}(x_i)$ as the the bias of model $h_t$ on sample $x_i$. 

To accelerate the convergence, we propose a knowledge transfer strategy, and $\beta$ controls the proportion of knowledge we should transfer from the former pre-trained neural network. After $T$ iterations, we can get $T$ base models. For the base model $h_t$, we can get its weight $\alpha_t$, and then add it to the ensemble model $H_t$.




\section{Efficient diversity-driven Ensemble}
In this section, we introduce our framework EDDE in detail. The overall framework is shown in Figure~\ref{Fig.Archi}. First, we describe the full pipeline and make an overview of EDDE. Next, we introduce the adaptive knowledge transfer strategy for fast training. Last, we explain our new diversity measure, diversity-driven loss and the Boosting-based framework.

\renewcommand{\algorithmicrequire}{\textbf{Input:}}
\renewcommand{\algorithmicensure}{\textbf{Output:}}

\begin{algorithm}
  \caption{EDDE}
  \label{alg1}
  \begin{algorithmic}[1]

  \REQUIRE 
    $ $\\
    Training set $D = {(x_1,y_1),(x_2,y_2),...,(x_m,y_m)}$;\\
    $y_i \in C, C = \{c_1,c_2,...,c_k\}$;\\
    $T$: Number of iterations;\\
    $\gamma$: Controls the strength of the diversity-driven loss;\\
    $\beta$: Controls the proportion of knowledge transfer;
    
  \ENSURE 
  $ $\\
    The boosted classifier:$H_T$
  \STATE $N = |D|$; $t =1$;
  \STATE $W_1(x_i)=1/N$;
  \STATE $h_1 \gets  I(D,W_1)$
  \STATE $\alpha_1=\frac{\sum_{i:h_1(x_i)= y_i}^{m} }{\sum_{i:h_1(x_i)\ne y_i}^{m}}$
  \STATE Update $H_1$ with $h_1$
  \FOR{$t=2$ to $T$}
  \STATE $h_{t} \gets  I(D,W_{t-1},h_{t-1},H_{t-1},\gamma,\beta)$
  \STATE $Sim_{t}(x_i) = 1 - \frac{\sqrt{2}}{2}\left \| \boldsymbol{h_{t}(x_i)} - \boldsymbol{H_{t-1}(x_i)}  \right \| $ 
  \STATE $Bias_{t}(x_i) = \frac{\sqrt{2}}{2}\left \| \boldsymbol{h_{t}(x_i)} - \boldsymbol{y_i}  \right \| $ 
  \STATE $W_{t}(x_i)=\frac{W_{1}(x_i)}{Z_t} e^{Sim_{t}(x_i) + Bias_{t}(x_i) }, h_t(x_i)\ne y_i$
  \STATE where $Z_t$  is a normalization factor 
  \STATE $\alpha_t=\frac{1}{2}log\frac{\sum_{i:h_t(x_i)= y_i}^{N}Sim_{t}(i)W_{t}(i) }{\sum_{i:h_t(x_i) \ne y_i}^{N}Sim_{t}(i)W_{t}(i)}$
  \STATE Update $H_t$ with $h_t$ and $\alpha_t$
 
  \STATE $t = t+1$
  \ENDFOR
  \STATE $H_T =  \sum_{i=1}^{T} \alpha_t h_t$
 
  \end{algorithmic}
  \label{A1}
  
\end{algorithm}

\subsection{Overview}
\begin{figure*}[htbp]
\centering
\includegraphics[width=13cm]{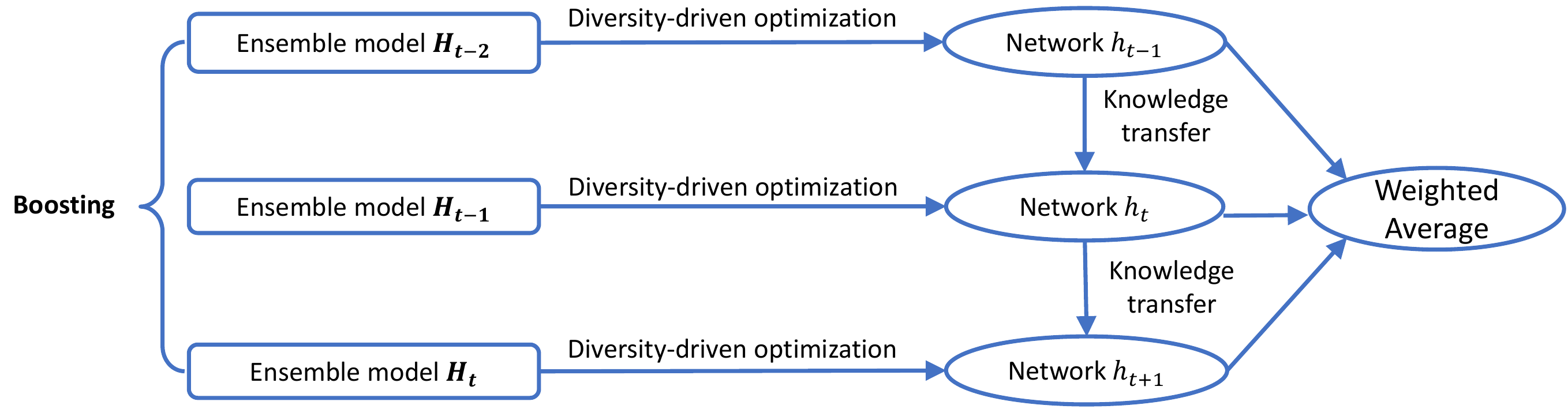}
\caption{Overview of EDDE}
\label{Fig.Archi}
\vspace{-12pt}
\end{figure*}

\noindent  To enhance diversity, we explicitly propose a diversity measure and use
a diversity-driven optimization method to train each base neural network. 
Besides, to accelerate the convergence, we propose a knowledge transfer strategy which can accelerate the training process without decreasing the diversity.
At last, we combine all the operations above in a Boosting-based framework. 
Figure~\ref{Fig.Archi} shows the full pipeline of EDDE.

Seen from Figure~\ref{Fig.Archi}, during the training process of base model $h_t$, we transfer the generic knowledge of the earlier fitted model $h_{t-1}$ to accelerate the convergence. Besides, we adopt a diversity-driven optimization to negatively correlate the soft-target of the former ensemble model $H_{t-1}$. After each iteration, we add the $h_t$ to the ensemble $H_{t-1}$ and get a new ensemble model $H_t$. Note that the weights of training samples are updated in each iteration and we use a Boosting-based strategy to construct the full pipeline.

The learning algorithm of EDDE is summarized in Algorithm~\ref{A1}, and all notations we used are defined in Table~\ref{Notation}. We will explain Algorithm~\ref{A1} in detail in the following parts.

\subsection{Adaptive knowledge transfer for fast training}
As the size of the ensemble grows, the training time grows linearly with the size of the ensemble. Therefore, it's desirable to find a method to accelerate the convergence of each base model.

A simple way to achieve this goal is to use Transfer Learning. Transferred with the previous knowledge, an additional network can be hatched from the former pre-trained neural network and trained significantly faster than trained from scratch. Practically, this allows us to train very large ensembles of deep neural networks in the time taken to train just a couple from scratch~\cite{Rapid}.

Transfer learning~\cite{pan2010survey} is a strategy where a machine learning model developed for a task is reused as the starting point for another model on a second task. In transfer Learning, we first train a base network on a base dataset, and then we repurpose the learned features, or transfer them, to a second target network to be trained on a target dataset.
We can naturally transfer the knowledge of previously trained neural network $h_{t-1}$ to train another neural network $h_t$, and $h_t$ may converge more quickly as it lies in a region near a good local minimum~\cite{seyyedsalehi2015fast}.

Take the Snapshot Ensemble for an example, all the weights of the pre-trained neural network $h_{t-1}$ are transferred to the current base neural network $h_t$. Indeed, we can rapidly get a converged model $h_t$.
However, the model $h_t$ may make similar predictions as $h_{t-1}$ because $h_t$ starts training at the local minimum proposed by $h_{t-1}$. That's why the diversity in Snapshot Ensemble is limited. Classically, transfer learning aims to transfer the information from one domain to another domain and it hasn't taken the diversity between each base models into consideration.

To guarantee the diversity in an ensemble, it is unsuitable to transfer all the pre-trained knowledge without selection. Fortunately, DNN generally learns generic features (e.g., edge detectors or color blob detectors) in its lower layers and task-specific features (e.g., parts of specific objects) in its upper layers [7]. The features in the upper layers are (weighted) combinations of the features in the lower layers and they are directly responsible for making predictions. Besides, the generic features contained in the earlier several layers of a neural network should be suitable for both base and target tasks, instead of specific to the base task.
Inspired by this discovery, we naturally come up with the idea of using these generic features to accelerate the training process of the next base model.

\begin{figure}[b]
\centering
\includegraphics[width=6cm]{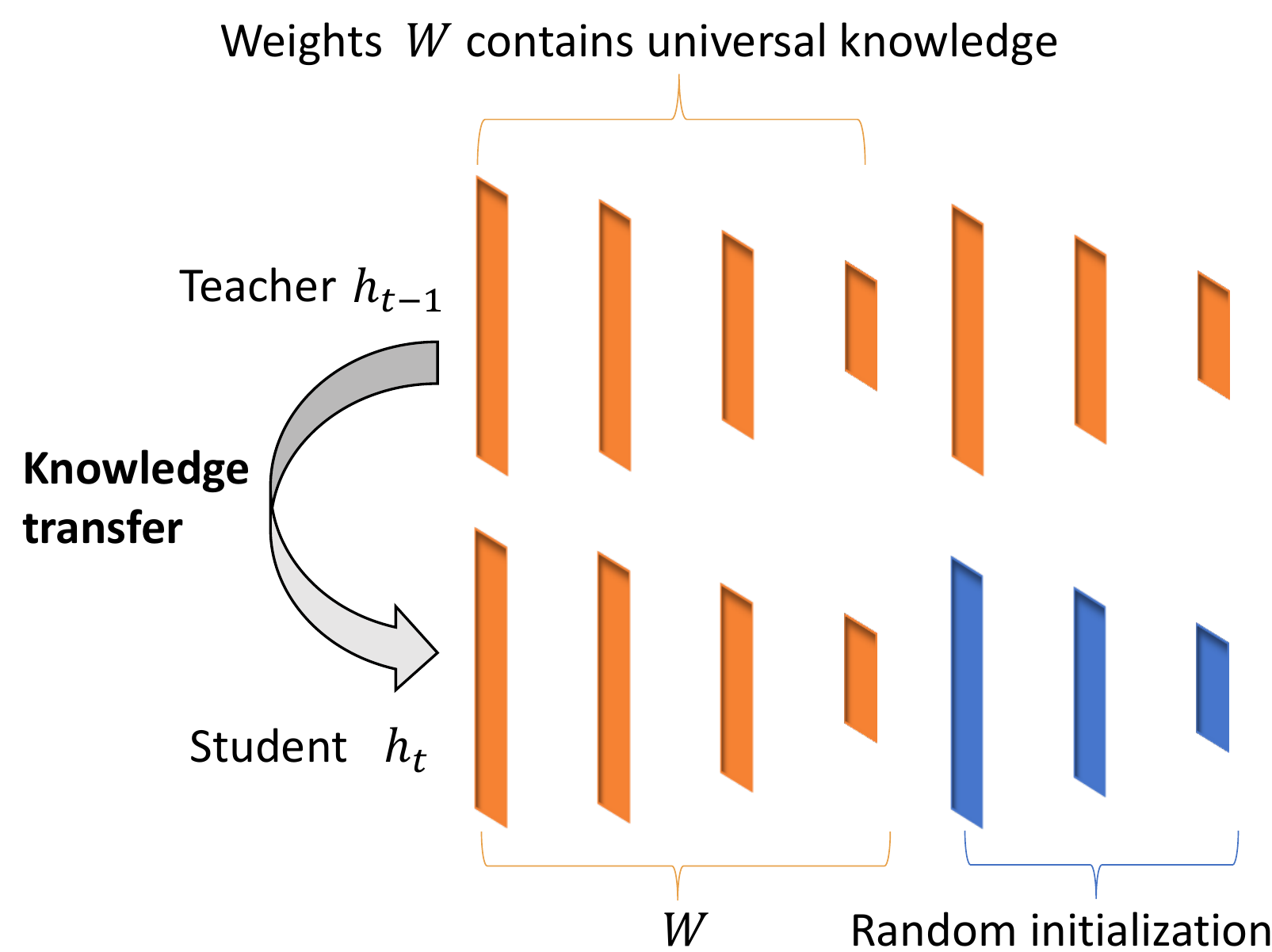}
\caption{The procedure of knowledge transfer}
\label{Fig.Initialization}
\end{figure}

The full procedure of our knowledge transfer is shown in Figure~\ref{Fig.Initialization}. We transfer the parameters of the earlier several layers of the teacher model $h_{t-1}$ to student model $h_t$.  
Besides, in order not to decrease the diversity, we randomly initialize $h_t$'s remaining high-level layers since they contain the specific knowledge of the given dataset.

Note that we fine-tune both the transferred wights of the low layers  and the randomly initialized weights of the high layers in each iteration, which means the transferred knowledge will not be fixed during the retraining processes. That's because the fixed transferred knowledge adds too strong constraints (i.e., most of the parameters are not allowed to be updated) to the optimization task, and a better local minimum is not likely to be
obtained given such constraints. By fine-tuning all the weights, each individual base model has the ability to converge to some good and different local minimums in a short time.

Like Snapshot Ensemble, we accelerate the convergence of each base model by transferring the knowledge of the former pre-trained neural network. The main difference is that we just transfer the generic knowledge.
The core problem we face is how to find such generic features, which is also equal to say how many layers of weights we should transfer from one pre-trained neural network to the other untrained one.

We propose an efficient and adaptive method to tackle this problem. In our proposed method EDDE, we denote the parameter $\beta$ to determine the proportion of parameters we should transfer. If it gets too much, the diversity may be reduced; if it is too small, we cannot maximize the training speed. Therefore, we aim to find a $\beta$ that leverages a trade-off between model diversity and training speed.



\begin{figure}[b]
\centering
\includegraphics[width=7.5cm]{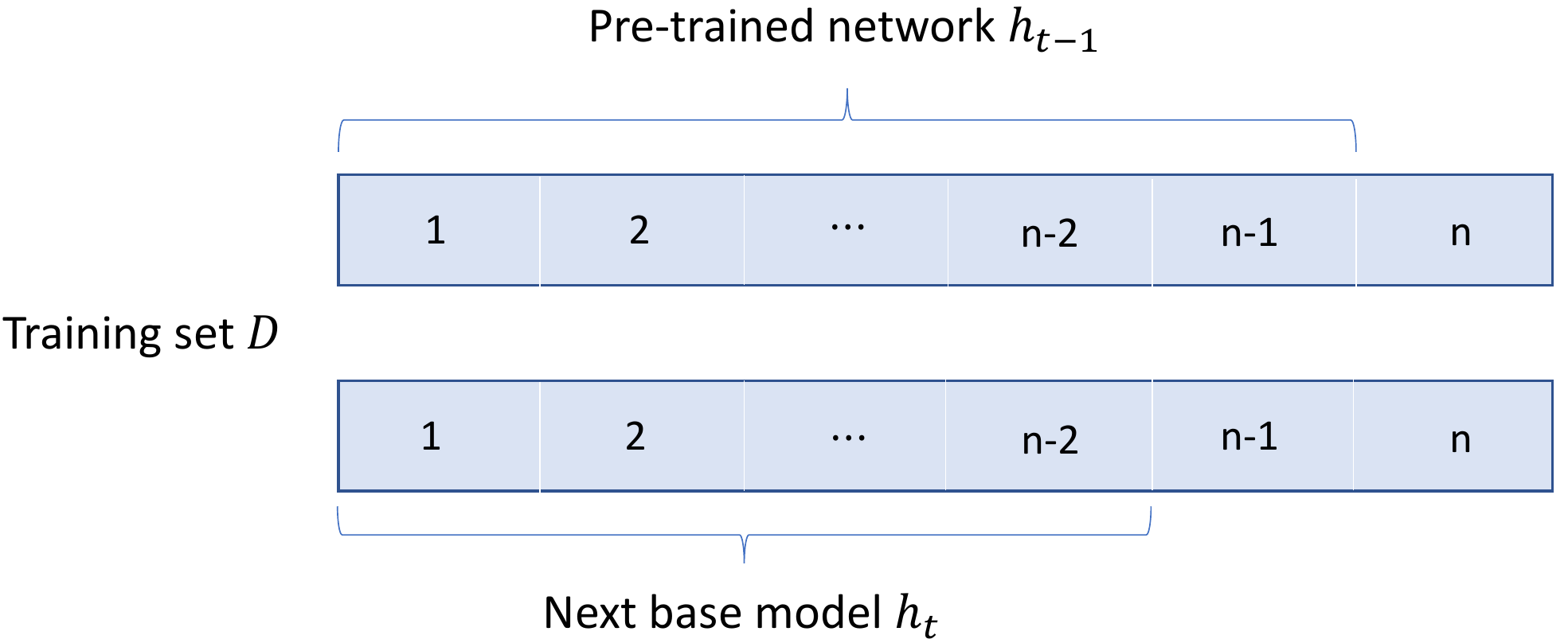}
\caption{The data split of knowledge transfer  }
\label{Fig.Tune}
\end{figure}

As shown in Figure~\ref{Fig.Tune}, to better analyze this problem, we assume
the training set $D$ is split into $n$ folds, and we leave the last fold as the test set.
We train a neural network $h_{t-1}$ using the first $n$-1 folds, and initialize the next network $h_{t}$ according to $\beta$.
Then, we train $h_{t}$ with the first $n$-2 folds.
Finally, we predict $h_{t}$'s accuracy on the ($n$-1)-th fold and the $n$-th fold.
In the above case, the pre-trained network $h_{t-1}$ has seen the ($n$-1)-th fold while the next base model $h_t$ hasn't. Besides, $h_{t-1}$ generally gets a higher accuracy on the ($n$-1)-th fold than the $n$-th fold due to overfitting. Different choices of $\beta$ can lead to different accuracy results:

\begin{figure}[htbp]
\centering
\includegraphics[width=8.8cm]{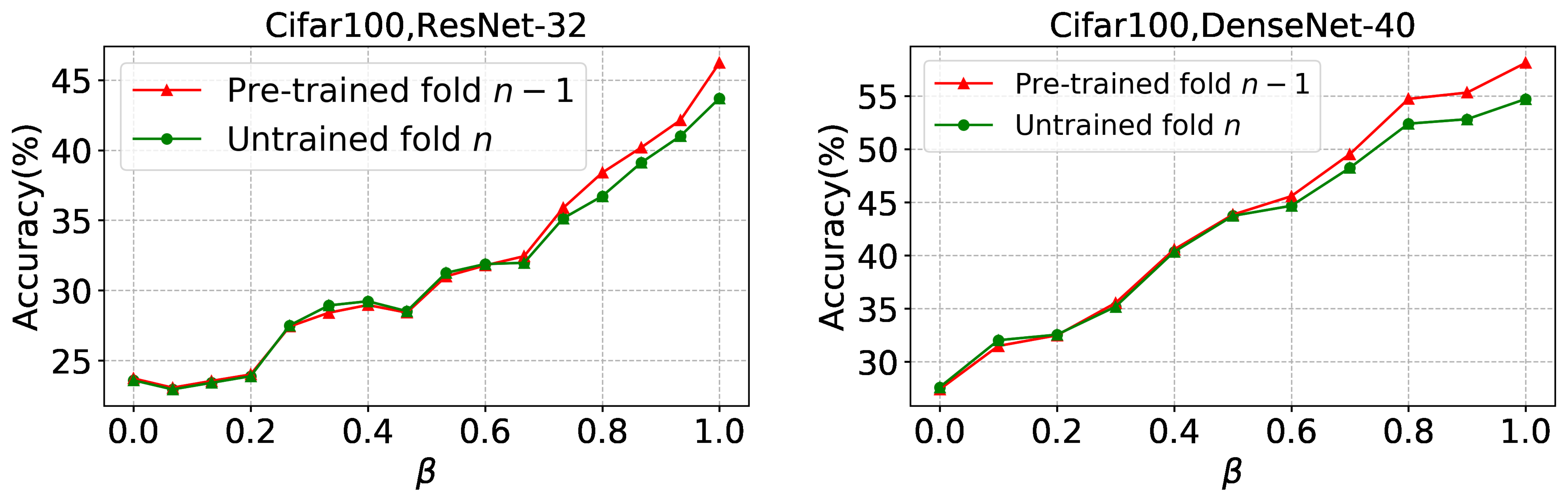}
\caption{The test accuracy using different parameter $\beta$}
\label{Fig.transfer}
\vspace{-10pt}
\end{figure}

\begin{itemize}
    \item 
    If $\beta$ is large, we transfer a large portion of weights of $h_{t-1}$, yielding a $h_t$ similar to $h_{t-1}$.
    Since $h_{t-1}$ has specific knowledge on the ($n$-1)-th fold, the accuracy of $h_t$ on the ($n$-1)-th fold is higher than that on the $n$-th fold.
    
    \item
    If $\beta$ is small, we transfer a small portion of weights of $h_{t-1}$, and therefore $h_t$ can be more different.
    Accordingly, $h_t$ has more possibility to forget the specific knowledge learned by $h_{t-1}$ from the ($n$-1)-th fold.
    Since $h_t$ has never seen both the ($n$-1)-th fold and the $n$-th fold, the accuracy on these two folds are similar.

\end{itemize}

Motivated by the above analysis, we find that we can choose a $\beta$ according to $h_t$'s performance on the dataset, which $h_{t-1}$ saw but $h_t$ did not.
In practice, we can start from $\beta=1$ and gradually reduce it, until $h_t$ performs similarly on two datasets --- an exclusive dataset owned by $h_{t-1}$ and a shared test dataset.
With this strategy, we can achieve a good trade-off between model diversity and training speed.

We also run an experiment to validate the above analysis. With different $\beta$, we split the training set CIFAR-100 into six folds ($n=6$). We firstly pre-train the model $h_1$ using the first 5 folds, then transfer the weights of $h_1$ to initialize our next base model $h_2$. During the training process of $h_2$, we calculate $h_2$'s mean accuracy of the first 5 epochs on the sixth fold and the fifth fold, the experimental result is shown in Figure~\ref{Fig.transfer}.  Seen from the experiments on ResNet-32 and DenseNet-40, as we gradually reduce $\beta$ to a suitable value, $h_2$'s accuracy on the fold $n-1$ may be similar to that on the untrained fold $n$.

Remarkably, the pre-trained model $h_{1}$ can still be used in the ensemble process. Besides, the parameter $\beta$ only needs to be tuned with the first base model $h_{1}$ and we just fix this tuned value for the latter base models.
In this way, we only need a few epochs to find how many parameters we need to transfer, which will not incur a large extra cost.

\subsection{Diversity measure}
\noindent  Various diversity measures have been proposed~\cite{mea} for the classifiers, and AdaBoost.NC is the first one to explore the ambiguity decomposition for classification ensembles. It names the penalty term as $amb$ and defines it with the correct/incorrect decision. Suppose the  practical output $h_{t}(x)$ is 1 if $x$ is labeled correctly, and -1 otherwise. Given an ensemble model $H$ and the weight $\alpha_t$, we get the definition
\begin{equation}
\small
amb=\frac{1}{2}\sum_{t=1}^{T}\alpha_t(H-h_t)
\label{amb}
\end{equation}

It's true that such a definition can be applied to a classification task, and can also be used in the negative correlated learning.
However, it has severe drawbacks in two aspects. On the one hand, they ignore the original softmax outputs of each base model and use the concrete classification label, such coarse-grained definitions may lose lots of useful information the model has learned before. On the other hand, as the ensemble output is the hard target, we cannot get useful gradient information with such a definition. As a result, it is impossible to directly optimize a diversity-driven loss function using this definition. That's also the reason why AdaBoost.NC enhance diversity just by changing the sample weights.

To sum up, it is desirable to propose a new diversity measure which can make full use of the softmax outputs of each network and can also be used in a diversity-driven optimization.

Therefore, we define a new diversity measure for two networks $h_j$ and $h_k$. The diversity between these two base models is denoted as
\begin{equation}
\small
Div_{h_j,h_k}= \frac{\sqrt{2}}{2}\frac{1}{N}\sum_{i=1}^{N}{\left \| \boldsymbol{h_{j}(x_i)} - \boldsymbol{h_{k}(x_i)} \right \|}_2
\label{Div_jk}
\end{equation}

Compared with the Eq.~\ref{amb}, our proposed diversity measure adopts the soft target, which contains more knowledge of the trained neural network. Besides, for each node in the softmax layer, we can get its gradient information with this definition. Such a feature is indispensable for our diversity-driven loss function.

Besides, we define the similarity between $h_j$ and $h_k$ as

\begin{equation}
\small
Sim_{h_j,h_k}= 1- Div_{h_j,h_k}
\label{Sim_jk}
\end{equation} 
The larger $Sim_{h_j,h_k}$ means model $h_j$ is more similar to model $h_k$, thus their diversity $Div_{h_j,h_k}$ becomes smaller accordingly. This definition will be used in our diversity-driven optimization latter.

As both the $\boldsymbol{h_{t}(x_i)}$  and $\boldsymbol{H(x_i)}$ are model's softmax outputs, we have:
\begin{equation}
\small
{\parallel \boldsymbol{h_{t}(x_i)} \parallel _1}=
{\parallel \boldsymbol{H(x_i)} \parallel _1}
= 1
\label{Norm0}
 \end{equation}

For a $k$-dimension vector $\boldsymbol{x}$, 
\begin{equation}
\small
 \parallel \boldsymbol{x} \parallel _2^2 = \sum_{i=1}^{k}{x_i}^2  \leq \sum_{i=1}^{k}{x_i}^2 + 2\sum_{i,j,i \neq j}^{k}\left|x_i \right|\left|x_j\right|= \parallel \boldsymbol{x} \parallel _1^2
 \label{Norm}
\end{equation}
According to Eq.~\ref{Norm0} and~\ref{Norm}, we get:
\begin{equation}
\small
 \begin{split}
 {\left \| \boldsymbol{h_{t}(x)} \!-\! \boldsymbol{H(x)} \right \|}_2^2 
   &\!=\! {\left \| \boldsymbol{h_{t}(x)} \right \|}_2^2 \!+\! {\left \| \boldsymbol{H(x)} \right \|}_2^2 \!-\! 2\boldsymbol{h_{t}}(x)\boldsymbol{H(x)}\\
   &\leq \parallel \boldsymbol{h_{t}(x)} \parallel _1^2+\parallel \boldsymbol{H(x)} \parallel _1^2 \\
   & -2\boldsymbol{h_{t}(x)}\boldsymbol{H(x)}\leq 2
 \end{split}
 \label{Norm2}
\end{equation}  

\begin{figure*}[htbp]
\centering
\includegraphics[width=11.5cm]{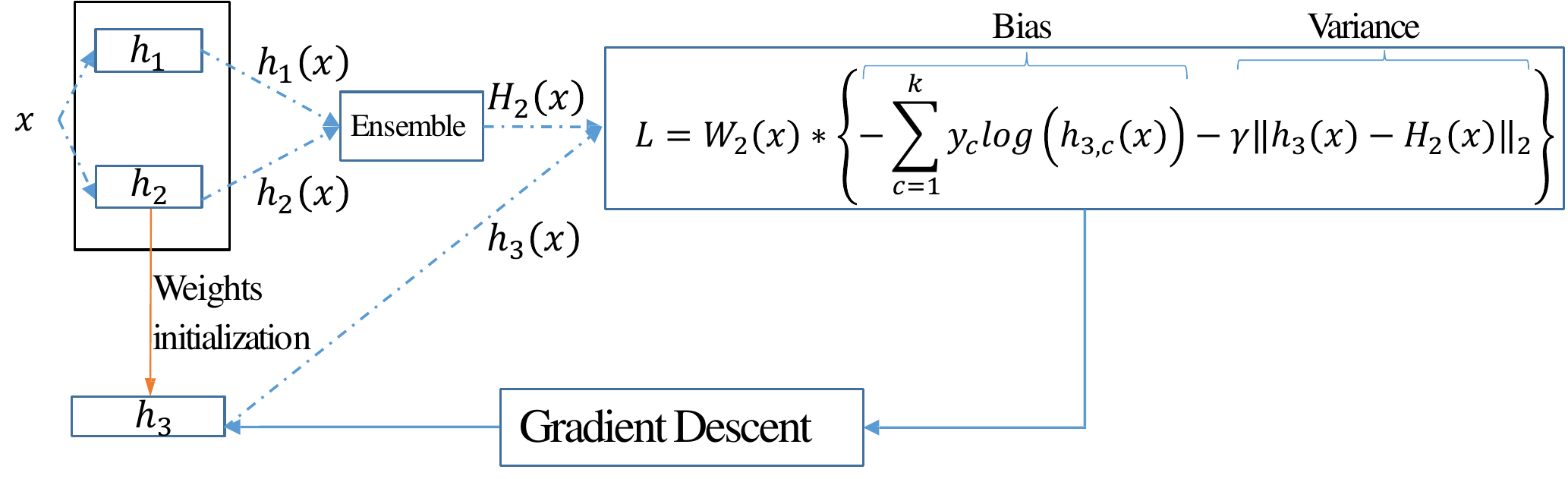}
\caption{The diversity-driven optimization of EDDE}
\label{Fig.Optimization}
\vspace{-10pt}
\end{figure*}

According to Eq.~\ref{Div_jk},~\ref{Sim_jk} and~\ref{Norm2}, both $Div_{h_j,h_k}$ and $Sim_{h_j,h_k}$ range in $\left[0,1\right]$. 

Last, to calculate the diversity of the ensemble network $H$, we define
\begin{equation}
\small
Div_{H}= \frac{2}{T(T-1)}\sum_{j=1}^{T}{\sum_{k=j+1}^{T}{Div_{h_j,h_k}}}
\label{Div_h}
\end{equation}

For numerically comparing the diversity of two ensemble model, we can use Eq.~\ref{Div_h} to calculate their $Div_H$ respectively.

\subsection{ Explicit diversity-driven loss for optimization}
According to Knowledge Distillation, a simple student model can be trained with the objective of matching the full softmax distribution of the complex teacher model. As the soft targets can provide much more information per training case and get much less variance in the gradient between training cases, a student network can achieve better accuracy by virtue of knowledge transferred from the teacher model than it would if trained directly~\cite{BANs}.

Motivated by the idea of Knowledge Distillation, EDDE also adopts the soft target in our loss function. However, the usage of the soft target is totally different in two aspects. First, we optimize the loss function with our self-defined diversity measure.
More importantly, the student network in Knowledge Distillation matches the soft target of the teacher model, but EDDE aims to negatively correlate this soft-target since our goal is not to enhance accuracy but to increase the difference between each base model and the ensemble model. Note that the student model in EDDE has the same architecture as the teacher model, so the accuracy of base models can be guaranteed in EDDE.

Specially, we define the ensemble teacher model by a function $y = H(x;w)$ where $x$ is the training sample, and $w$ is the learned weights of $H$. Both the $H(x_i;w)$ and  $h(x_i;w')$ are the vector of probabilities representing the conditional distribution over object categories given sample $x_i$.
One of our objectives is to find a new set of parameters $w'$ for a student  neural network $h(x;w')$ such that
\begin{equation}
\small
w'= \mathop{\arg\max}_{w'} \sum_{i=1}^{N}\ \ \| h(x_i;w') - H(x_i;w)\|_2
\label{arg}
\end{equation} 

According to Eq.~\ref{arg}, the base model $h(x;w')$ in each iteration is more likely to make different prediction to the former ensemble model $H(x_i;w)$.

Based on our diversity measure and the above objective, we define the penalty term as

\begin{equation}
\small
amb=\gamma\left \| \boldsymbol{h_{t}(x)} - \boldsymbol{H_{t-1}(x)} \right \|_2
 \end{equation}
 
Suppose we use the categorical cross-entropy as the loss function.
We have two objectives when designing the objective function. 
The first objective is to minimize the error (Bias in Figure~\ref{Fig.Optimization}) of the base models, which is the cross entropy with the correct labels.
The other is to maximize their difference (Variance in Figure~\ref{Fig.Optimization}).
Based on this intuition, we define a weighted diversity-driven loss function on sample $x$ as

\begin{equation}
\small
\resizebox{.9\hsize}{!}{
 $L\!=\!W_{t-1}(x)\!\bigg\{\!-\!\sum_{c= 1}^{k}y_clog(h_{t,c}(x))\!-\!\gamma\left \| \boldsymbol{h_{t}(x)}-\! \boldsymbol{H_{t-1}(x)} \right \|\!_2\bigg\}\!$}
\end{equation}  

The hyperparameter $\gamma$ is used to adjust the strength of the second objective. Larger $\gamma$ may introduce higher diversity, but the bias of the neural network may also be increased accordingly. As a result, this parameter needs to be carefully tuned in our experiments.

It is easy to get the gradient information with our diversity-driven loss function. For $h_t$'s output node $c$, the partial derivative of $L$ with respect to the output $h_{t,c}$ is 

\begin{equation}
\small
\resizebox{.9\hsize}{!}{
 $\frac{\partial L}{\partial h_{t,c}(x)}\!=\!W_{t-1}(x)\bigg\{ \!-\!\frac{y_c}{h_{t,c}(x)}\!-\!\gamma\frac{h_{t,c}(x)\!-\!H_{t-1,c}(x)}{\left \| \boldsymbol{h_{t}(x)} \!-\! \boldsymbol{H_{t-1}(x)} \right \|_2}\bigg\}$}
 \label{BP}
\end{equation}

We use the standard back propagation (BP) algorithm ~\cite{Rumelhart1988Learning} to optimize each network and its weights are updated with Eq.~\ref{BP}.
To explain the optimization process in detail, we take the ensemble of three base models for an example.
As shown in Figure~\ref{Fig.Optimization}, if we pre-train two base models $h_1$ and $h_2$, we can combine them and get the ensemble model $H_2$. Given the training data $D$ with sample weights $W_2$, the pre-trained knowledge by $h_2$, and two parameters $\gamma$ and $\beta$, we train the third model $h_3$ to negatively correlate the softmax outputs of $H_2$, such procedure is correspond to the $I(D,W_2,h_2,H_2,\gamma,\beta)$ in line 7 of algorithm ~\ref{A1}.


\subsection{ Boosting-based framework}

To further enhance diversity, we use a Boosting-based framework to combine the diversity-driven optimization and the knowledge transfer strategy. As an art of combining the predictions of different machine learning models, Boosting is a very popular ensemble method and it has been widely used in many deep learning scenes~\cite{BoostedCNN}.
It increases the weight of the misclassified samples for the next training process, as the weight of each sample can be changed in each cycle and the loss function is closely related to the sample weights, the optimization path is changed correspondingly and each base model may make different predictions as they can converge to different local minimums.

For each sample $x_i$, we set $Sim_{t}(x_i)$ as the similarity between model $h_t$ and $H_{t-1}$ on sample $x_i$ (line 8 of Algorithm~\ref{A1}). Besides, we set $Bias_{t}(x_i)$ as the bias of model $h_t$ on sample $x_i$ (line 9 of Algorithm~\ref{A1})

\begin{equation}
\small
 \begin{split}
 Sim_{t}(x_i) = 1-\frac{\sqrt{2}}{2}\left \| \boldsymbol{h_{t}(x_i)} - \boldsymbol{H_{t-1}(x_i)} \right \|_2
 \end{split}
 \label{Sim_t}
\end{equation}  

\begin{equation}
\small
Bias_{t}(x_i) = \frac{\sqrt{2}}{2}\left \| \boldsymbol{h_{t}(x_i)} - \boldsymbol{y_i}  \right \|_2
\label{Bias}
\end{equation}  




For a misclassified sample $x_i$, we update its weights by (line 10 of Algorithm~\ref{A1})
\begin{equation}
\small
W_{t}(x_i)=\frac{W_{1}(x_i)}{Z_t} e^{Sim_{t}(x_i) + Bias_{t}(x_i)}
\end{equation} 

Larger $Sim_{t}(x_i)$ means the individual base model $h_t$ has the same opinion with the ensemble model $H_{t-1}$ on sample $x_i$. If sample $x_i$ is mis-classified by $h_t$, the ensemble model $H_{t-1}$ probably cannot classify this sample either, thus we must give more attention on sample $x_i$ and the weight $W_{t}(x_i)$ will be increased more. On the contrary, if $Sim_{t}(x_i)$ is small, which means $H_{t-1}$ may has better ability to classify the sample $x_i$, as it's prediction is less similar to $h_t$. In this way, it's better to increase the $W_{t}(x_i)$ less aggressively.

Note that the base models of traditional Boosting algorithm only needs to be weak and it is true that we can get a strong ensemble model if we have a large number of weak base learners. However, it's hard to meet such condition in deep learning as each base model needs a high training cost.
So for EDDE, we hope the base models have both high diversity and accuracy. As the sample weights we used for these base models are only to enhance diversity, we just update the weights based on $W_1$. Due to this operation, the weight of each sample can be greatly changed. Since the loss function is closely related to the sample weights, the optimization path is changed correspondingly and the diversity can be further increased.

At last, we define the weight of each base model as (line 12 of Algorithm~\ref{A1})
\begin{equation}
\small
\alpha_t=\frac{1}{2}log\frac{\sum_{i:h_t(x_i) = y_i}^{N}Sim_{t}(x_i)W_{t}(x_i) }{\sum_{i:h_t(x_i) \ne y_i}^{N}Sim_{t}(x_i)W_{t}(x_i)}
\end{equation}

After we get the weight $W_{t}(x_i)$ and the similarity $Sim_{t}(x_i)$ of each sample $x_i$, we can accordingly calculate the weight of each base model. A large $Sim_{t}(x_i)$ means $h_t(x_i)$ is very similar to $H_{t-1}(x_i)$, which indicates the model $h_t$ has the same opinion with the ensemble model $H_{t-1}$. Usually, the prediction of the ensemble model $H_{t-1}$ is relatively more accurate, thus $h_t$'s prediction becomes more important accordingly. In this way, we can increase or decrease $\alpha_t$ more if the $Sim_{t}(x_i)$ is large.

For the prediction process, we average the softmax outputs of each base model $h_t$ with the model weight $\alpha_t$, and the ensemble model $H_T$ is defined as (line 16 of Algorithm~\ref{A1})
\begin{equation}
\small
H_T =  \sum_{i=1}^{T} \alpha_t h_t
\end{equation} 


Previous works have shown that training neural network ensemble through sub-sampled dataset may lead to low generalization accuracy as it reduces the number of unique data items seen by an individual neural network. A neural network has a large number of parameters and it is affected relatively more from this reduction in unique data items than the ensemble of other classifiers such as decision trees or SVMs~\cite{why}.
Therefore, we use all the training set in each iteration.

To sum up, our method differs from traditional Boosting algorithms in three aspects. 
\begin{itemize}
    \item we use all the training data instead of using the sub-sampling technique in each iteration as the less unique training samples may lead to low bias. 
    \item We train each base models based on the knowledge of the pre-trained one while the traditional Boosting methods train every network from scratch and introduce a high training cost. 
    \item we use a diversity-driven loss function to get each base model, and the weights update we use is totally different from the traditional Boosting methods. 
\end{itemize}

\section{Experiments}
To prove the validity and efficiency of EDDE, we conduct extensive experiments on the task of CV and NLP. We firstly introduce our experimental settings in section V-A and then study the effectiveness of our method EDDE in Section V-B. To clearly show the high efficiency of EDDE, we make an end-to-end comparison with other ensemble methods in Section V-C. In Section V-D, we show how diversity influences the generalization ability of an ensemble model. Besides, we simply analyze the influence of the parameter in Section V-E. At last, to prove the effectiveness of our diversity-driven loss and the knowledge transfer strategy, we make an ablation study in Section V-F. All the programs were implemented in Python using the Keras library$\footnote{https://github.com/keras-team/keras}$.

\subsection{Experimental settings}
\paragraph{Datasets} 
We use the The CIFAR-10 (C10) and CIFAR-100 (C100) datasets~\cite{Cifar} for the CV task. Besides, we we also test EDDE in the NLP tasks with the IMDB~\cite{IMDB} and MR~\cite{pang} datasets.

The CIFAR-10 (C10) and CIFAR-100 (C100) datasets consist of 60000 32x32 pixels color images, and consist of 10 and 100 classes respectively. 50,000 images were used for training and another 10,000 ones were used for testing on each dataset. A widely used data augmentation scheme~\cite{he2016deep} is used before the training process. 

Both the IMDB and MR dataset are  reviews with one sentence per review, and they have been labeled as positive or negative.  
For the preprocessing of IMDB, we set the max length of each sentence to 120 and the max features to 5000. Note that the max length means we cut texts after this number of words and max features mean we just use the top max features most common words. As for the MR dataset, we use the same settings in~\cite{kim2014convolutional}.

\paragraph{Base networks} 
For the CV task, we train a Residual Neural Network with 32 layers (ResNet-32) and a Densely Connected Convolutional Networks with 40 layers (DenseNet-40)~\cite{DenseNet}, and the growth rate of DenseNet we use is 12. 
For the NLP task, we use the Text-CNN~\cite{kim2014convolutional} as the base model.


\paragraph{Baselines} 
To show the improvement of the ensemble learning, we firstly compare our method with a \textbf{Single Model}. 
Besides, we compared EDDE with \textbf{BANs}~\cite{BANs} as we are both motivated by the idea of Knowledge Distillation and we both use the soft target.
To validate the importance of diversity, we compare with \textbf{Snapshot Ensemble}~\cite{huang2017snapshot} as it introduces low diversity even it can accelerate the training process. 
Besides, as our method \textbf{EDDE} is a diversity-driven framework for neural networks, we also compare with the diversity-driven ensemble method \textbf{AdaBoost.NC}~\cite{wang2010negative}.
Last, we also run the experiments with \textbf{Bagging}~\cite{breiman1996stacked} and \textbf{AdaBoost.M1}~\cite{Ada} to validate the limitations of traditional ensemble methods in the field of deep learning.

\paragraph{Protocol}
We train these networks with the stochastic gradient descent and set the initial learning rate to 0.1 for ResNet and Text-CNN, and 0.2 for DenseNet. Besides, All methods except Snapshot Ensemble use a standard learning rate schedule that we divide the learning rate by 10 when the training is at 50\% and 75\% of the total training epochs. 
For EDDE using ResNet, we set the parameters $\gamma$ to 0.1 and $\beta$ to 0.7. As for the DenseNet, $\gamma$ and $\beta$ are 0.2 and 0.5. For EDDE in NLP task, we transfer the knowledge of all the convolution layers of Text-CNN to initialize the next base model.
For Snapshot Ensemble, we use the same settings as the original paper. The mini-batch we use for the MR, CIFAR and IMDB datasets are 50, 64 and 128 respectively. 

\begin{figure*}[htbp]
\centering
\includegraphics[width=\textwidth]{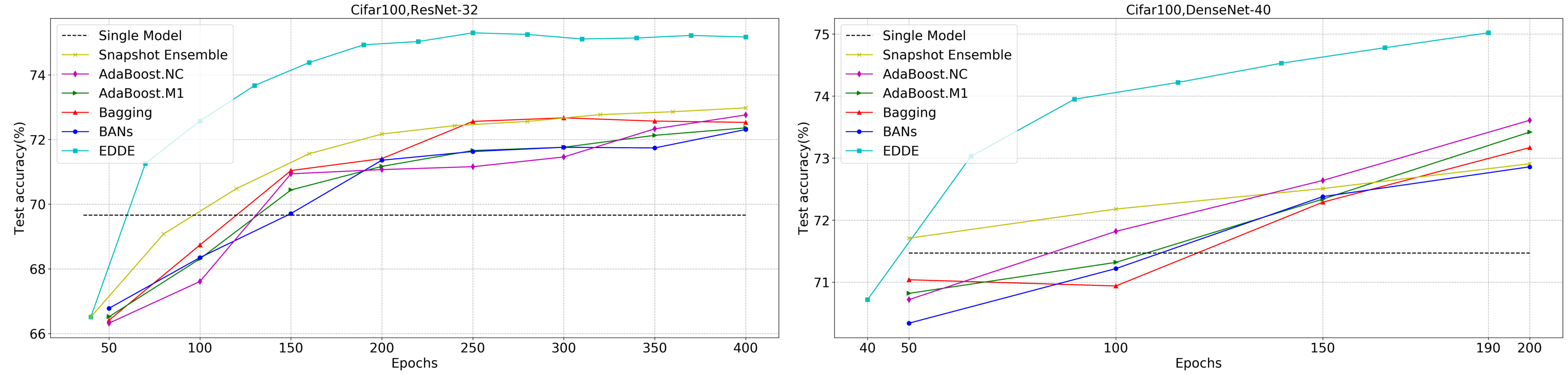}
\caption{Test accuracy of different ensemble methods on CIFAR-100 using ResNet-32 (Left) and DenseNet-40 (Right). For the Single Model, the test accuracy is directly calculated on the test set in the last epoch. For the ensemble method, the test accuracy is the ensemble accuracy which is calculated with the base models already trained.}
\label{Fig.C100}
\end{figure*}

\begin{figure*}[htbp]
\centering
\includegraphics[width=15cm]{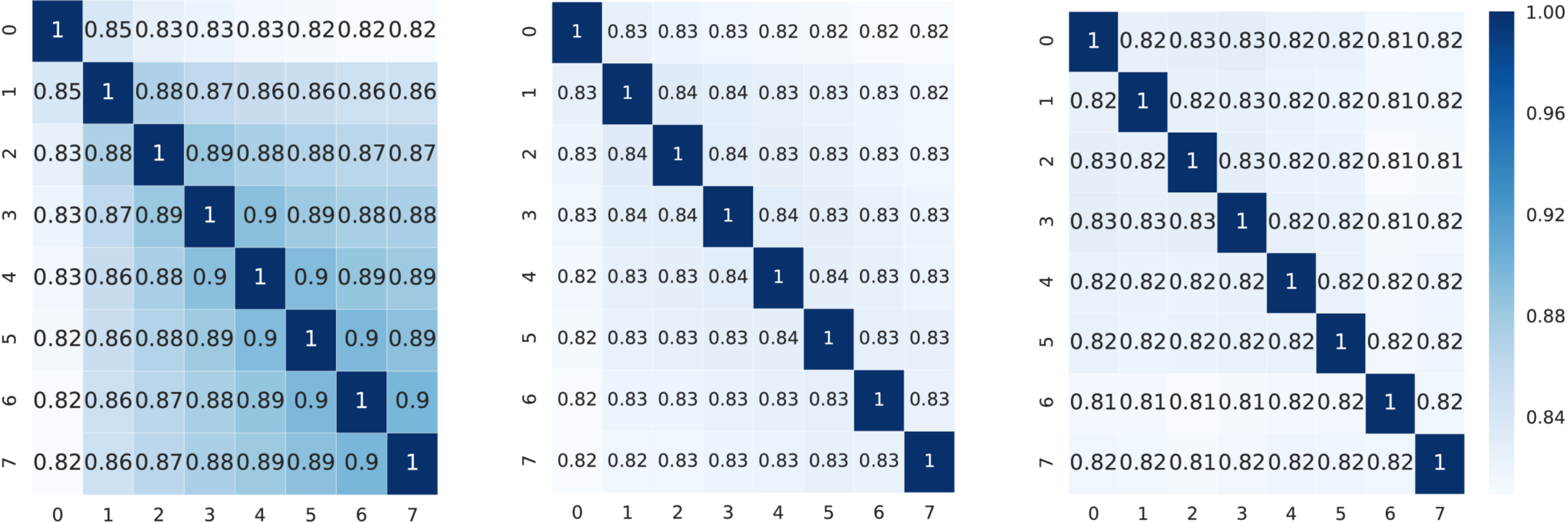}
\caption{The pairwise similarity between Snapshot Ensemble(left), EDDE(middle), and AdaBoost.NC(right) }
\label{Fig.Diversity}
\vspace{-8pt}
\end{figure*}

\paragraph{Training budget.} 
For Bagging, AdaBoost.M1, AdaBoost.NC, and BANs, each base model is trained with a budget of 50 epochs for CIFAR dataset and 20 epochs for IMDB dataset. 
For Snapshot Ensemble on CIFAR dataset, we use the same settings in~\cite{huang2017snapshot}, in which snapshot variants are trained with 4 cycles (50 epochs per cycle) for DenseNets and 10 cycles (40 epochs per cycle) for ResNets. As for the IMDB and MR datasets, snapshot variants are trained with  5 cycles (20 epochs per cycle).
For EDDE, we train the first base model with the same settings in Snapshot Ensemble, and each cycle after that is trained with 30 epochs for ResNet, 25 epochs for DenseNet and 10 epochs for Text-CNN.

\subsection{Effectiveness study}

\begin{table}[!htbp]
\caption{Test accuracy on the CV task} 
\centering
{
\noindent
\renewcommand{\multirowsetup}{\centering}
\begin{tabular}{cccc}
\toprule
 \textbf{Model} &\textbf{Method}& \textbf{C10}&\textbf{C100}\\
\midrule
\multirow{8}*{ResNet-32}
 & Single Model&  92.73\% & 69.11\%\\
 & BANs& 92.81\%&  71.36\%\\
 & Bagging& 92.58\%&  71.41\% \\
 & AdaBoost.M1& 92.22\%&   71.17\%\\\
 & AdaBoost.NC& 92.64\%&   71.07\%\\
 & Snapshot& 93.27\%&  72.17\%\\
 & EDDE& \textbf{\blue{94.11\%}}& \textbf{\blue{74.38\%}}\\
\midrule
\multirow{6}*{DenseNet-40}
 &Single Model& 92.61\%& 71.47\%\\
 & BANs& 93.11\%& 72.86\%\\
 & Bagging&93.24\%& 73.17\%\\
 & AdaBoost.M1&92.87\%& 73.42\%\\
 & AdaBoost.NC&93.17\%& 73.61\%\\
 & Snapshot& 92.91\%& 72.91\%\\
 & EDDE&  \textbf{{\blue{94.39\%}}}&  \textbf{{\blue{75.02\%}}}\\
\bottomrule
\end{tabular}}
\label{Fig.acc1}
\vspace{-10pt}
\end{table}

\begin{table}[!htbp]
\caption{Test accuracy on the NLP task} 
\centering
{
\noindent
\renewcommand{\multirowsetup}{\centering}
\begin{tabular}{cccc}
\toprule
 \textbf{Model} &\textbf{Method}& \textbf{IMDB}&\textbf{MR}\\
\midrule
\multirow{8}*{Text-CNN}
 & Single Model&  86.61\% &  76.14\%\\
 & BANs& 86.98\%&  76.23\%\\
 & Bagging& 87.14\% &  76.51\%\\
 & AdaBoost.M1& 86.72\%&  76.17\%\\\
 & AdaBoost.NC& 86.87\%&  76.26\%\\
 & Snapshot& 86.91\%&  76.43\%\\
 & EDDE& \textbf{{\blue{87.69\%}}}& \textbf{{\blue{76.98\%}}}\\
\bottomrule
\end{tabular}}
\label{Fig.acc2}
\vspace{-12pt}
\end{table}

To validate the effectiveness of EDDE, we train different ensemble methods on different tasks, and the main results are summarized in Table~\ref{Fig.acc1} and Table~\ref{Fig.acc2}. For  CIFAR dataset, the methods in the same group are trained for 200 epochs. EDDE is trained for 50 epochs for the IMDB and MR dataset, and the other methods in the same group are trained for 100 epochs. Note that the results of our method are colored in \blue{blue}, and the best result for each network/dataset pair is bolded.

For the CV task, the experiments on CIFAR data show that EDDE can always get the highest ensemble accuracy in all cases. Take the result on CIFAR-100 using ResNet-32 for an example, EDDE achieves the accuracy rate of 74.38\%, far outperforming the next-best model's 72.17\% under the same training cost.
For the NLP task, EDDE only need half time to achieve 87.69\% test accuracy in IMDB dataset and 76.98\% test accuracy in the MR dataset, which means EDDE is superior the other ensemble methods both in the speed and accuracy.
Through the experiments on the CV and NLP tasks, we observe that EDDE is capable of getting a high ensemble accuracy with a limited training budget.

\subsection{End-to-end comparison} 
As the most concern of EDDE is to efficiently get a high ensemble accuracy with a limited training budget, it's necessary to make an end-to-end comparison to other ensemble methods. 
For EDDE and other baselines, we train each base model with the same network structures and dataset. Therefore, the training time per epoch for each ensemble method is same.
Besides, the extra time cost brought by different ensemble method is trivial compared to the training time of deep neural networks. 
As a result, the training epochs can be treated as the training expense. 
We compare the ensemble accuracy of each method trained with different epochs in the CV tasks, and the result is shown in Figure~\ref{Fig.C100}.

We firstly see that the AdaBoost.M1, AdaBoost.NC and Bagging all have relatively low test accuracy during each period. The main reason is that they train each base model individually on a sub-sampled dataset, thus it's hard to get enough base models with high accuracy in a short time. For BANs, it also gets a low ensemble accuracy in many cases. That is because it randomly initialized each base model without the prior knowledge and it cannot get high diversity and accuracy at the same time.

As we can see in Figure~\ref{Fig.C100},  EDDE gets a higher test accuracy than other methods in all cases. More concretely, from the left picture, we see EDDE achieves the accuracy of 73.67\% within only 130 epochs, while the next-best model Snapshot Ensemble needs 400 epochs to achieves 72.98\% ensemble accuracy, which means EDDE is more than 3 times faster than the next-best model in this scenario. 

Similarly, as shown in the right picture in Figure~\ref{Fig.C100}, EDDE always gets higher ensemble accuracy than other methods using DenseNet-40. Because our knowledge transfer method is capable of accelerating the training process and the diversity-driven method can explicitly enhance diversity, EDDE can get the highest efficiency.

\begin{table*}[!htbp]
\caption{Compare the influence of diversity}
\centering
{
\noindent
\renewcommand{\multirowsetup}{\centering}
\begin{tabular}{ccccccc}
\toprule
\textbf{Method}& 
 \textbf{Training epochs}&
 \textbf{Average accuracy}& \textbf{Ensemble accuracy}&
 \textbf{Increased accuracy}& \textbf{Diversity}\\
\midrule
Snapshot Ensemble&400&68.53\%&72.98\%& 4.45\%&0.1322 \\

EDDE&250&68.04\%&\textbf{{\blue{75.30\%}}}&7.26\%&0.1702\\
AdaBoost.NC&400&66.81\%&72.76\%& 5.95\%&0.1787 \\
\bottomrule
\end{tabular}}
 \label{C100_result}
\end{table*}

\subsection{Diversity analysis} 
We run an experiment on CIFAR100 using ResNet-32 and analyze the diversity between their first 8 base models. According to Eq.~\ref{Sim_jk}, we compute the pairwise similarity among each base model of Snapshot Ensemble, EDDE, and AdaBoost.NC. The result is shown in Figure ~\ref{Fig.Diversity}. 

For Snapshot Ensemble, the similarity between two nearby base models is high and it becomes higher when the model is trained longer. That's because the next model is initialized with the former one, it's more prone to converge to nearby local minimums during the two nearby training cycle. Besides, if trained longer, the base models becomes more accurate and they have more chance to make similar predictions. 

On the contrary, it's clear to see that the pairwise similarity of EDDE and AdaBoost.NC is smaller than Snapshot Ensemble. To further explore the significance of the diversity, we compute the increased accuracy of each ensemble methods. Besides, we also compute their concrete diversity according to Eq.~\ref{Div_h}. The experimental results are summarized in Table~\ref{C100_result}.


As shown in Table~\ref{C100_result}, the AdaBoost.NC can get the highest diversity of 0.1787. That's because it trains the base models on different sub-sampled datasets. Besides, it randomly initializes the weights of each base model and then negatively correlates the former neural network by adjusting the sample weights. However, even given 400 training epochs, it gets the lowest average accuracy of 66.81\%, which means all the base model cannot converge to a good local minimum given limited time.

As for the Snapshot Ensemble, it gets the highest average accuracy of 68.53\% among these three methods since it transfers the knowledge of the pre-trained neural network to accelerate the training process of the current model. Unfortunately, as it transfers all the knowledge without selection, it gets the lowest diversity. 

Seen from Table~\ref{C100_result}, the base models in EDDE can get both high diversity and average accuracy.
Accordingly, EDDE gets the highest increased accuracy of 7.26\% . Besides, EDDE achieves the 75.30\% accuracy only using 250 epochs, while Snapshot Ensemble and AdaBoost.NC cannot get such accuracy even if trained with 400 epochs. Therefore, EDDE is efficient in the ensemble of deep neural networks.





\subsection{Exploration of hyperparameters}
In order to investigate the influences of hyperparameters on EDDE, we vary their values and compare the ensemble accuracy accordingly.
EDDE has two hyperparameters, $\gamma$ controls the strength of the diversity-driven loss and $\beta$ determines the proportion of knowledge we should transfer from the pre-trained network. As we introduced in Section IV-B, we have proposed an effective method to find the optimal $\beta$, so the only parameter we should tune in EDDE is $\gamma$. Therefore, we vary the parameter $\gamma$ and run the experiment on ResNet-32 using CIFAR 100 dataset. The result is summarized in Table~\ref{para}. 


\begin{table}[!htbp]
\vspace{-8pt}
\caption{Test accuracy with different parameter} 
\centering
{
\noindent
\renewcommand{\multirowsetup}{\centering}
\begin{tabular}{ccc}
\toprule
 \textbf{Method} &\textbf{Parameter}& \textbf{Ensemble accuracy}\\
\midrule
\multirow{5}*{EDDE}
 & $\gamma$ = 0&  73.86\% \\
 & $\gamma$ = 0.1& \textbf{{74.38\%}}\\
 & $\gamma$ = 0.3& 74.13\% \\
 & $\gamma$ = 0.5& 73.72\%\\\
 & $\gamma$ = 1& 72.47\%\\
\bottomrule
\end{tabular}}
\label{para}
\end{table}

For analyzing the influence of $\gamma$, we set $\gamma$ to 0, 0.1,0.3,0.5 and 1. We can observe from Table~\ref{para} that EDDE with the setting $\gamma$ of 0.1 get the highest accuracy of 74.38\% since EDDE can get the most balanced tradeoff between diversity and accuracy in this setting. 
However, when setting $\gamma$ to 0, the ensemble accuracy is decreased to 73.86\%. In this situation, our diversity-driven loss function is equal to the normal loss function, and the diversity of EDDE is reduced accordingly. 
On the contrary, if we set $\gamma$ to 1, the ensemble accuracy has a sharp decline. A higher $\gamma$ means we negatively correlate the previous soft target more, and it also means we give less attention to the true target. Accordingly, the network cannot converge well with a very high $\gamma$. However, according to Table~\ref{Fig.acc1}, even set $\gamma$ to 1, EDDE trained with 200 epochs still outperforms Snapshot Ensemble trained with 400 epochs. Therefore, our method EDDE is robust to the hyperparameters.

\subsection{Ablation study}
 In order to show the effectiveness of our diversity-driven loss function and the knowledge transfer strategy, we add the ablation studies to measure their effect respectively. 

We firstly train an ensemble model using EDDE without diversity-driven loss function, this strategy is denoted as EDDE (normal loss). Besides, like Snapshot Ensemble, we transfer all the pre-trained knowledge in EDDE, and we name this method EDDE (transfer all). Next, we individually train each base model without knowledge transfer, and this method is denoted as EDDE (transfer None). At last, we also compare EDDE with AdaBoost.NC using transfer learning (initialize each model using the weights of the pre-trained one), we name this method AdaBoost.NC (transfer). All these methods are tested in the  CIFAR100 dataset using ResNet-32 and the training budget is 200 epochs for EDDE and 400 epochs for Adaboost.NC . The experimental results are summarized in Table~\ref{Abl}. 


\begin{table}[!htbp]
\vspace{-8pt}
\caption{ablation study} 
\centering
{
\noindent
\renewcommand{\multirowsetup}{\centering}
\resizebox{88mm}{12mm}{
\begin{tabular}{cccc}
\toprule
\textbf{Method}& \textbf{Ensemble accuracy}& \textbf{Diversity}& \textbf{Average accuracy}\\
\midrule
EDDE&  \textbf{\blue{74.38\%}}&  0.1743&67.91\%\\
EDDE (normal loss)&   73.86\%&   0.1682&67.97\%\\
EDDE (transfer all)&   73.37\%&   0.1631&68.16\%\\
EDDE (transfer none)&  70.78\%& 0.1854&66.72\%\\
AdaBoost.NC (transfer)&   72.64\%&  0.1573&67.33\%\\
\bottomrule
\end{tabular}}}
\label{Abl}
\end{table}

As shown in Tabel~\ref{Abl}, the ensemble accuracy may decrease if we use a normal loss function. Besides, we can get a higher average accuracy of 68.16\% if we transfer all the pre-trained knowledge, but the diversity may decrease accordingly. 
On the contrary, from the result of EDDE (Transfer none), we can get the highest diversity of 0.1854. However, given limited training budget, each base model cannot converge well without the knowledge transfer. As a result, EDDE (Transfer none) gets the lowest average accuracy and ensemble accuracy.

As for the AdaBoost.NC with transfer learning, it can get higher average accuracy compared with the original AdaBoost.NC, but its ensemble accuracy is not as high as EDDE due to its lower diversity and average accuracy. That's because it trains base models on the sub-sampled datasets and transfers all the previous knowledge without selection.

 Among all these methods, EDDE can get the most balanced tradeoff between average accuracy and diversity and it can get the highest ensemble accuracy of 74.38\%, thus
both the diversity-driven loss and the knowledge transfer are effective in EDDE.

\section{Conclusion}
Ensemble learning is useful in improving the generalization ability of deep neural networks. However, given limited training budget, current ensemble methods are not efficient as they cannot balance the tradeoff between diversity and accuracy. We proposed EDDE to tackle this problem. 
To improve diversity, we proposed a new diversity measure and optimize each base model with a diversity-driven loss function.
To enhance the training speed and improve the accuracy, we proposed a knowledge transfer method which can efficiently transfer the generic knowledge from the pre-trained model.
Last, we adopt a Boosting-based framework to further improve diversity and combine the operations above.
Experimental results on the CV and NLP tasks have shown that our method EDDE is efficient in generating multiple neural networks with high diversity and accuracy. Given the same training budget, EDDE can get a more accurate ensemble model compared with the baselines.

\section{Acknowledgement}
This work is supported by the National Key Research and Development Program of China (No. 2018YFB1004403), NSFC (No. 61832001, 61702015, 61702016, 61572039), and PKU-Tencent joint research Lab.

\end{document}